\documentclass[10pt,twocolumn,letterpaper]{article}

\usepackage[pagenumbers]{cvpr} % To force page numbers, e.g. for an arXiv version

\usepackage{graphicx}
\usepackage{amsmath}
\usepackage{amssymb}
\usepackage{booktabs}
 
\usepackage[utf8]{inputenc} % allow utf-8 input
\usepackage[T1]{fontenc}    % use 8-bit T1 fonts
\usepackage{url}            % simple URL typesetting
\usepackage{booktabs}       % professional-quality tables
\usepackage{amsfonts}       % blackboard math symbols
\usepackage{nicefrac}       % compact symbols for 1/2, etc.
\usepackage{microtype}      % microtypography
\usepackage[export]{adjustbox}

\usepackage{bm}
\usepackage{xcolor}
\usepackage{amsmath}
\usepackage{common}
\usepackage{caption}
\usepackage{subcaption}
\usepackage{multirow}
\usepackage{algorithm}
\usepackage{algpseudocode}
\usepackage{nicefrac}
\usepackage{graphics}
\usepackage{xspace}
\usepackage{tabularx}

\usepackage{amssymb}% http://ctan.org/pkg/amssymb
\usepackage{pifont}% http://ctan.org/pkg/pifont
\newcommand{\cmark}{\ding{51}}%
\newcommand{\xmark}{\ding{55}}%

\usepackage{amsmath,amsfonts,bm}

\def\eqref#1{equation~\ref{#1}}
\def\1{\bm{1}}

\def\vc{{\bm{c}}}

\def\vh{{\bm{h}}}

\def\vm{{\bm{m}}}

\def\vw{{\bm{w}}}
\def\vx{{\bm{x}}}
\def\vy{{\bm{y}}}

\def\mW{{\bm{W}}}

\DeclareMathAlphabet{\mathsfit}{\encodingdefault}{\sfdefault}{m}{sl}
\SetMathAlphabet{\mathsfit}{bold}{\encodingdefault}{\sfdefault}{bx}{n}

\DeclareMathOperator*{\argmax}{arg\,max}
\DeclareMathOperator*{\argmin}{arg\,min}

\usepackage[pagebackref,breaklinks,colorlinks]{hyperref}

\usepackage[capitalize]{cleveref}
\crefname{section}{Sec.}{Secs.}
\Crefname{section}{Section}{Sections}
\Crefname{table}{Table}{Tables}
\crefname{table}{Tab.}{Tabs.}

\def\cvprPaperID{*****} % *** Enter the CVPR Paper ID here
\def\confName{CVPR}
\def\confYear{2022}

\newcommand{\mpar}{{\mathbf{\theta}}}
\newcommand{\exm}{f^{ExM}}

\newcommand{\exml}{ExML\xspace}
\newcommand{\exalgo}{{\cal{A}}}

\newcommand{\lexp}{{\mathit{e}}}
\newcommand{\Dorig}{{\cal{D}}}

\newcommand{\Dex}{{\cal{D}}^{ex}}
\newcommand{\buffer}{{\cal{M}}}
\newcommand{\synmem}{{\cal{M}}^{ex}}
\newcommand{\synmemsub}{{\tilde{\cal{M}}}^{ex}}

\newcommand{\classexp}{{\cal{Y}}}

\newcommand{\fexpert}{f^{S}}
\newcommand{\fexm}{f^{ExM}}

\newcommand{\loss}{{\cal{L}}}

\newcommand{\real}{\mathbb{R}}

\newcommand{\emptycell}{ \multicolumn{1}{c}{--} }

\begin{document}

\title{Ex-Model: Continual Learning from a Stream of Trained Models}

\author{Antonio Carta\\
University of Pisa\\
{\tt\small antonio.carta@di.unipi.it}
\and
Andrea Cossu\\
Scuola Normale Superiore\\
{\tt\small andrea.cossu@sns.it}
\and
Vincenzo Lomonaco\\
University of Pisa\\
{\tt\small vincenzo.lomonaco@unipi.it}
\and
Davide Bacciu\\
University of Pisa\\
{\tt\small davide.bacciu@unipi.it}
}
\maketitle

%%%%%%%%% ABSTRACT
\begin{abstract}
  Learning continually from non-stationary data streams is a challenging research topic of growing popularity in the last few years. Being able to learn, adapt, and generalize continually in an efficient, effective, and scalable way is fundamental for a sustainable development of Artificial Intelligent systems. 
  However, an agent-centric view of continual learning requires learning directly from raw data, which limits the interaction between independent agents, the efficiency, and the privacy of current approaches. Instead, we argue that continual learning systems should exploit the availability of compressed information in the form of trained models. In this paper, we introduce and formalize a new paradigm named \emph{"Ex-Model Continual Learning"} (\exml), where an agent learns from a sequence of previously trained models instead of raw data. We further contribute with three ex-model continual learning algorithms and an empirical setting comprising three datasets (MNIST, CIFAR-10 and CORe50), and eight scenarios, where the proposed algorithms are extensively tested. Finally, we highlight the peculiarities of the ex-model paradigm and we point out interesting future research directions.
  %We provide a number of continual learning algorithms and, through empirical results based on three different benchmarks, eight scenarios, and an extensive study of the proposed algorithm, we show the peculiarities of this paradigm as well as pointing out interesting future research directions.
\end{abstract}

%%%%%%%%% BODY TEXT
%%%%%%%%%%%%%%%%%%%%%%%%%%%%%%%%%%%%%%%%%%%%%%%%%%%%%%%%%%%%%%%%%%%%%
\section{Introduction}
%%%%%%%%%%%%%%%%%%%%%%%%%%%%%%%%%%%%%%%%%%%%%%%%%%%%%%%%%%%%%%%%%%%%%

\begin{figure*}[th]
    \centering
      \includegraphics[width=0.95\textwidth]{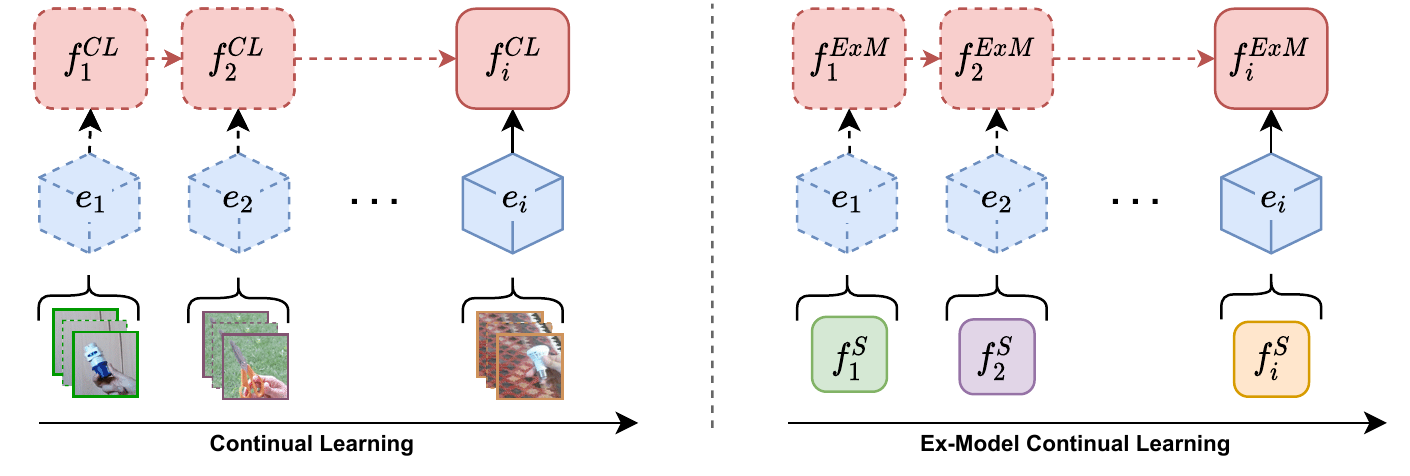}
      \captionof{figure}{Ex-model Continual Learning. The stream is composed of expert models, without access to the original data.}
      \label{fig:exmodel}
\end{figure*}

\emph{Continual learning} (CL) studies learning in dynamic, non-stationary environments \cite{parisiContinualLifelongLearning2019,masanaClassincrementalLearningSurvey2020}. Recently, there has been significant progress in the development of continual learning algorithms able to efficiently learn deep hierarchical representations from a sequence of experiences or tasks with increasingly robust and effective solutions, even for challenging scenarios with high degrees of non-stationarity \cite{lesortContinualLearningRobotics2019,masanaClassincrementalLearningSurvey2020}. 

Most of these solutions follow an agent-centric view of Artificial Intelligence, which tends to mimic the same operative constraints of biological learning systems  \cite{ringCHILDFirstStep1997,lesortContinualLearningRobotics2019,hayesReplayDeepLearning2021}. Under this view, a continual learning agent directly interacts with the environment and learns from \emph{raw data}. This framework is closer to neuroscience-grounded theories of learning and intelligence \cite{hayesReplayDeepLearning2021,kemkerFearNetBrainInspiredModel2018}, but it ignores the opportunities and challenges provided by the pervasive and distributed nature of the modern computing infrastructure: 
\begin{enumerate}
    \setlength{\itemsep}{0pt}
    \setlength{\parskip}{0pt}
    \setlength{\parsep}{0pt}
    \item \emph{expert models}: continual learning should reuse knowledge from expert models, such as local personalized models or large pretrained models.
    \item \emph{distributed learning}: agents in a distributed environment should be able to learn independently and to share knowledge efficiently at the same time.
    \item \emph{sample efficiency}: learning from raw data may be inefficient due to noise and redundancy inherent to high-dimensional perceptual data.
    \item \emph{privacy}: sharing knowledge between agents must be limited by privacy constraints, and each agent should be allowed to set its privacy constraints.
\end{enumerate}
Currently, (1) is partially addressed by initializing continual learning models using pretrained models \cite{hayesLifelongMachineLearning2020a,maltoniContinuousLearningSingleincrementaltask2019}. However, it is not possible to use multiple pretrained models or to exploit a pretrained model after the initialization phase.
Recently, some works have partially addressed (2) by studying federated continual learning \cite{yoonFederatedContinualLearning2021,usmanovaDistillationbasedApproachIntegrating2021,hendryxFederatedReconnaissanceEfficient2021}. Unfortunately, this approach requires a tight integration between the devices, intensive communication and strong assumptions about the model's architecture and learning procedure. Point (3) is often ignored in the continual learning literature. Pretrained models can partially address (3) by providing a compressed form of knowledge, i.e. the model's parameters, that can be used to learn more efficiently. Lastly, (4) is explored in CL with settings such as data-free class-incremental scenarios \cite{smithAlwaysBeDreaming2021, choiDualTeacherClassIncrementalLearning2021}, where access to the previous data is forbidden. Again, this scenario assumes a single agent and access to the current data, making it difficult to share knowledge between multiple agents. 

In this paper, we propose a novel framework based on an alternative and integrative approach of the four points above, envisioning a more pervasive and distributed form of continual learning. Compressed knowledge and skills in the form of trained neural models (\emph{"neural skills"}, for short) are generated and made available every day. So our motivating question is rather: \emph{why not to learn directly from them instead?} Learning directly from trained models allows to efficiently share knowledge between domain experts (1), to train each expert independently (2) and efficiently (3). Additionally, each expert can determine its privacy level (4) by not sharing the model or training with differentially private algorithms \cite{abadiDeepLearningDifferential2016}. 

Intuitively, learning from models resembles other forms of learning from compressed knowledge, such as when we learn from books or use the Internet instead of learning by trial and error. We argue that learning from compressed knowledge will become more and more important for the same reasons. 
Towards this vision, the original contributions of this paper can be summarized as follows:
\begin{enumerate}
    \setlength{\itemsep}{6pt}
    \setlength{\parskip}{6pt}
    \setlength{\parsep}{0pt}
    \item We propose and formalize \emph{Ex-Model Continual Learning}\footnote{The meaning of \emph{ex} comes from latin, which can be roughly translated as "out of, from". } (\exml) as a new CL paradigm designed to allow efficient and private sharing of compressed knowledge between independent agents (Section \ref{sec:problem_formulation}).
    \item We propose a family of continual learning strategies, \emph{Ex-Model Distillation (ED)}, based on data-free knowledge distillation (Section \ref{sec:ex_model_distill}). In particular, we compare three possible instances of Ex-Model Distillation: two of them perform distillation by generating synthetic data, while the other relies on out-of-distribution data unrelated to the task solved by the expert (Section \ref{sec:datagen}).
    \item We assess the performance of Ex-Model Distillation strategies against five different baselines, three popular continual learning benchmarks (MNIST, CIFAR-10 and CORe50) and scenarios (Task-Incremental, Domain-Incremental and Class-Incremental) in order to highlight the general applicability of our solutions (Section \ref{sec:experiments}). We release the code to easily instantiate the \exml scenario and to reproduce our experiments\footnote{\url{https://github.com/AntonioCarta/ex_model_cl}}.
\end{enumerate}

We believe that the introduction of the \exml paradigm, together with the design of the Ex-Model Distillation strategies and the experimental setup, will provide a robust starting point to study continual learning from models and its impacts on many downstream applications (Section \ref{sec:discussions}).

%%%%%%%%%%%%%%%%%%%%%%%%%%%%%%%%%%%%%%%%%%%%%%%%%%%%%%%%%%%%%%%%%%%%%
\section{Ex-Model Continual Learning}\label{sec:problem_formulation}
%%%%%%%%%%%%%%%%%%%%%%%%%%%%%%%%%%%%%%%%%%%%%%%%%%%%%%%%%%%%%%%%%%%%%
We begin by formalizing a {\it classic} continual learning scenario (Figure \ref{fig:exmodel}), where data arrives in a streaming fashion as a (possibly infinite) sequence of learning experiences $S = \lexp_1, \hdots, \lexp_n$. We assume a supervised classification problem, where each experience $e_i$ consists of a batch of samples $\Dorig^i$, where each sample is a tuple $\langle x^i_k, y^i_k\rangle$ of input and target, respectively, and the labels $y^i_k$ are from the set ${\cal{Y}}^i$, which is a subset of the entire universe of classes ${\cal{Y}}$. Notice that it is very easy to generalize the scenario to different CL problems. Usually $\Dorig^i$ is split into a separate train set $\Dorig^i_{train}$ and test set $\Dorig^i_{test}$. A continual learning algorithm~ $\exalgo^{CL}$ is a function with the following signature\cite{lesortContinualLearningRobotics2019}:
\begin{equation}
    \exalgo^{CL}:\ \langle f^{CL}_{i-1}, \Dorig^i_{train}, \buffer_{i-1}, t_i\rangle\ \rightarrow\ \langle f^{CL}_i, \buffer_{i}\rangle
\end{equation}
where $f^{CL}_i$ is the model learned after training on experience $\lexp_i$, $\buffer_i$ a buffer of past knowledge, such as previous samples or activations, stored from the previous experiences and usually of fixed size. The term $t_i$ is a task label which may be used to identify the correct data distribution. Most of the experiments in this paper assumes the most challenging scenario of $t_i$ being unavailable. 
Usually, CL algorithms are limited in the amount of resources that they can use, and they are designed to scale up to a large number of training experiences without increasing their computational cost over time. 
The objective of a CL algorithm is to minimize the loss $\loss_{S}$ over the entire stream of data $S$:
\begin{align}
    \loss_{S}(f^{CL}_n, n) = \frac{1}{\sum\limits_{i=1}^n |\Dorig_{test}^i|} \sum_{i=1}^n \label{eq:cl_objective} \loss_{exp}(f^{CL}_n, \Dorig^i_{test}) \\
    \loss_{exp}(f^{CL}_n, \Dorig^i_{test}) = \sum_{j=1}^{|\Dorig_{test}^i|} \loss(f^{CL}_n(\vx^i_j), y^i_j),
\end{align}
where the loss $\loss(f^{CL}_n(\vx), y)$ is computed on a single sample $\langle\vx, y\rangle$, such as cross-entropy in classification problems. 

\paragraph{Stream of Experts} In an \exml scenario, there is \emph{no direct access to a data stream} $\Dorig_1, \hdots, \Dorig_n$. Instead, the stream consists of expert models $\fexpert_1, \hdots, \fexpert_n$, where each expert is trained on some specific domain (Figure \ref{fig:exmodel}). As a consequence, an ex-model algorithm must learn only by extracting information from each expert. To keep the scenario as general as possible, we do not make any assumption about the models, such as their architecture or the specific hyperparameters used during training. We assume that the \exml algorithms have no control over how the experts have been trained. Each model $\fexpert_i$ has been trained on a corresponding learning experience $\lexp_i$ to minimize $\loss_{exp}(\fexpert_i, \Dorig^i_{train})$. We denote by $\fexpert_i$ the learned function, $\mpar^{S}_i$ its parameters, $\fexpert_i(\vx)$ the model's output, i.e. the logits, for a particular input sample $\vx$, and $p^{S}_i(\vx)$ the output probabilities computed by applying the softmax function to the model's output.

\paragraph{ExML Scenario} The objective of the \exml scenario is to continuously update a model $\fexm_i$ whenever a new expert $\fexpert_i$ becomes available. Notice that the loss $\loss_{exp}(\fexm_i, \Dorig^i_{train})$ cannot be evaluated since we do not have access to the original data. Since the stream of models may be unbounded, training strategies must be scalable up to a large number of experts. Therefore, ex-model algorithms cannot keep in memory all the previous experts. As a result, there are two constraints in an \exml scenario: \emph{lack of access to the original data and limited computational resources}. 

Overall, an \exml algorithm $\exalgo^{ExM}$ is a function with the following signature:
\begin{equation}
    \exalgo^{ExM}: \langle\fexm_{i-1}, \fexpert_{i}, \synmem_{i-1}, t_i\rangle \rightarrow \langle\fexm_i, \synmem_{i}\rangle,
\end{equation}
where $\fexm_{i}$ is the current model, $\fexpert_{i}$ the current expert from the stream, $\synmem_{i-1}$ is a set of samples from out-of-distribution data or synthetically generated and currently available to the model (Section \ref{sec:datagen}), and $t_i$ the task label information. Again, notice that task labels are optional and they may not be available in many scenarios.
The objective of ex-model algorithms is to minimize Eq. \ref{eq:cl_objective}, the loss over the original (and unavailable) data stream.

%%%%%%%%%%%%%%%%%%%%%%%%%%%%%%%%%%%%%%%%%%%%%%%%%%%%%%%%%%%%%%%%%%%%%
\section{Ex-Model Distillation}\label{sec:ex_model_distill}
%%%%%%%%%%%%%%%%%%%%%%%%%%%%%%%%%%%%%%%%%%%%%%%%%%%%%%%%%%%%%%%%%%%%%

In this paper, we propose a family of algorithms, called Ex-model Distillation (ED) algorithms, to solve ex-model continual learning. The core idea behind our strategy is to exploit a cumulative buffer of synthetic or auxiliary data, generated from the expert model, to train the ex-model using knowledge distillation~\cite{hintonDistillingKnowledgeNeural2015}. In this section, we describe how to perform the distillation and defer the data generation process to Section \ref{sec:datagen}. The algorithm consists of two steps: buffer update and knowledge distillation. An overview of the algorithm is provided in Algorithm~\ref{algo:ex_distill}.

\paragraph{Buffer Update} Let us assume to have access to a set of samples $\synmem_{i-1}$ of fixed size $N$, where samples $\langle\vx^{syn}, y^{syn}\rangle$ are obtained from the previous steps of the algorithm and they act as surrogate data in place of the original data from $\lexp_1, \hdots, \lexp_{i-1}$. We use a data generating procedure $\exalgo^{gen}$ to generate a new set of samples
\begin{equation}
    \Dex_i = \exalgo^{gen}(\fexpert_i, \frac{N}{i}),
\end{equation}
where $|\Dex_i| = \frac{N}{i}$. $\exalgo^{gen}$ generates synthetic data using the expert $\fexpert_i$. To obtain a new buffer $\synmem_{i}$ of size $N$, we subsample $\synmemsub_{i-1} = subsample(\synmem_{i-1})$, such that $|\synmemsub_{i-1}| = N - \frac{N}{i}$ and combine it with the new data to obtain the updated buffer $\synmem_i = \synmemsub_{i-1} \cup \Dex_i$.

\paragraph{Knowledge Distillation} Once we have updated the synthetic buffer, we can start the distillation process. Differently from knowledge distillation, we need to distill knowledge from two different models, the previous ex-model $\fexm_{i-1}$, and the current expert from the stream $\fexpert_i$. Ex-model algorithms use $\synmem_i$ to distill the knowledge from the current expert without forgetting previous knowledge. Each of these models is trained on a different (possibly overlapping) set of classes: $\classexp^{prev} = \bigcup_{k=0}^{i-1} \classexp^{k}$ for the ex-model, and $\classexp^{i}$ for the expert. Given a sample $\langle\vx^{syn}, y^{syn}\rangle$ and the output $\vy^{curr} = \fexm_i(\vx^{syn})$ from the current ex-model, the target logits $\tilde{\vy}$  are computed by combining the normalized logits of the previous ex-model and current expert:
\begin{align}
    &\vy^{ExM} = normalize(\fexm_{i-1}(\vx^{syn})) \\
    &\vy^{S} = normalize(\fexpert_i(\vx^{syn})) \\
    &\tilde{\vy} = \begin{cases}
        \vy^{ExM}                 & \textit{if}\ y^{syn} \in \classexp^{prev}\\
        \vy^{S}                   & \textit{if}\ y^{syn} \in \classexp^{i}\\
        \frac{1}{2}(\vy^{ExM} + \vy^{ExM})     & \textit{if}\ y^{syn} \in \classexp^{prev} \wedge y^{syn} \in \classexp^{i}. \label{eq:y_target}
    \end{cases}
\end{align}
Output normalization allows to combine the outputs independently from the difference in scale between the two models, which would create a bias if not removed. The resulting vector $\tilde{\vy}$ is used as a target for the distillation by minimizing the Mean Squared Error (MSE) loss
\begin{equation}
    \loss_{MSE}(\vy^{curr}, \tilde{\vy}) = \norm{\vy^{curr} - \tilde{\vy}}^2_2. \label{eq:mse_distillation}
\end{equation}
Eq. \ref{eq:mse_distillation} by itself is not sufficient to train a good model. The main limitation of the loss is that it is unable to distinguish whether a sample $\vx^{syn}$ should be classified by the previous ex-model (i.e., it belongs to one of the previous experiences) or by the current expert (i.e. it belongs to the current experience) since the units of each model are treated separately. Therefore, the ex-model distillation loss $\loss_{ED}$ combines the MSE with the crossentropy loss $\loss_{CE}$:
\begin{align}
\begin{split}
    \loss_{ED}(\vy^{curr}, \tilde{\vy}, y^{syn}) =  &\loss_{MSE}(\vy^{curr}, \tilde{\vy}) \\
    + &\lambda_{CE} \loss_{CE} (\vy^{curr}, y^{syn}). \label{eq:ed_loss}
\end{split}
\end{align}
The loss in Eq. \ref{eq:ed_loss} is optimized by stochastic gradient descent for a fixed number of iterations, sampling randomly the buffer at each step to create a mini-batch.
\begin{algorithm} \label{alg:ed}
    \centering
    \caption{Ex-Model Distillation}\label{algo:ex_distill}
    \begin{algorithmic}[1]
        \Require Stream of pretrained experts $S$ and a continually learned model $\fexm$.
        \State $\synmem_0 \gets \{ \}$  \Comment{empty buffer}
        \For{$\fexpert_i$ in $S$}
            \State $\Dex_i \gets \exalgo^{gen}(\fexpert_i, \frac{N}{i})$
            \State $\synmemsub_{i-1} \gets subsample(\synmem_{i-1})$
            \State $\synmem_i \gets \synmemsub_{i-1} \cup \Dex_i$
            \For {$k$ in $1, \hdots, n_{iter}$}  \Comment{Knowledge Distillation}
                \State $\langle\vx^k, \vy^k\rangle \gets \textit{sample}(\synmem_i)$
                \State $\vy^{curr} \gets \fexm(\vx^k)$
                \State $\tilde{y} \gets \textit{get\_target}(\vx^k)$ \Comment{Eq. \ref{eq:y_target}}
                \State $L \gets  \loss_{ED}(\vy^k, \tilde{\vy}^k, y^k)$
                \State do SGD step on $L$
            \EndFor
        \EndFor
    \end{algorithmic}
\end{algorithm}

%%%%%%%%%%%%%%%%%%%%%%%%%%%%%%%%%%%%%%%%%%%%%%%%%%%%%%%%%%%%%%%%%%%%%
\section{Distillation Data}\label{sec:datagen}
%%%%%%%%%%%%%%%%%%%%%%%%%%%%%%%%%%%%%%%%%%%%%%%%%%%%%%%%%%%%%%%%%%%%%
As discussed in Section \ref{sec:problem_formulation}, ex-model distillation needs an alternative source of data $\synmem_i$ to distill the knowledge from the expert. Since the original data $\Dorig_i$ is not available, we need an alternative source of samples. In this section, we show three methods that can be used to generate a synthetic dataset. Notice that in order to obtain a good performance it is not necessary that the synthetic data resembles the original data. Even highly distorted images or images from different domains may be useful to distill the knowledge from $\exm_i$. In fact, we will see in the experimental results that the synthetic data that we will use may be widely different from the original data. %We evaluate three different methods to obtain distilled data.

\paragraph{Model Inversion.} Model inversion extracts samples using $\fexpert_i$ by maximizing the output probabilities of the chosen class by gradient descent. Given a randomly initialized sample $\vx^{syn}$, a target class $y^{syn}$, and an expert $\fexpert_i$, with model inversion we optimize $\vx^{syn}$ by stochastic gradient descent by minimizing the crossentropy $\loss_{CE} \left(\nicefrac{\fexpert_i(\vx^{syn})}{\tau}, y^{syn} \right)$, where $\tau$ is the softmax temperature. We can generate a batch of images for each class by using different random initializations. %Notice that since the computations for each sample are independent, they can be optimized in parallel using large mini-batches.
Since the computations for each sample are independent, they can be optimized in parallel using large mini-batches.

\paragraph{Data Impression.} Data impression is a data extraction method proposed in \cite{nayakZeroShotKnowledgeDistillation2019} for the data-free offline training scenario. Data Impression exploits the classifier's weights of the expert $\mW \in \real^{N_c \times N_h}$, where $N_h$ is the number of hidden units and $N_c$ the number of classes, to define a Dirichlet distribution used to sample probability targets. Data impression treats each row $\vw_k$ of $\mW$ as a template for class $k$, computing the matrix of pairwise similarities $C(k, j) = \frac{\vw_k^\top \vw_j}{\norm{\vw_k}, \norm{\vw_j}}$. The similarity coefficients are used to define a Dirichlet distribution $Dir(N_c, \vct{\alpha}^k)$ for each class such that $\vct{\alpha}^k = \beta \vc_k$, $\vy^{di} \sim Dir(N_c, \vct{\alpha}^k)$, where $\beta$ is a temperature parameter and $\vc_k$ the $k$th row of the similarity matrix. Targets $\vy^{di}$ sampled from the resulting Dirichlet distribution are used to optimize a randomly initialized $\vx^{syn}$ using a knowledge distillation loss $\loss_{KD}(p_{\exm_i}(\vx^{syn}), \vy^{di}, \tau)$.
We generate a different target for each sample using the Dirichlet distribution corresponding to the desired class.  Since data impression provides a target for the entire output distribution instead of a single target class, which is needed for the ex-model distillation loss (Eq. \ref{eq:ed_loss}), we set $y^{syn} = \argmax \vy^{di}$. The advantage of Data Impression compared to Model Inversion is that the Dirichlet distribution of the soft targets models the class similarities instead of ignoring them.

\paragraph{Auxiliary Data.} The usage of auxiliary data is an alternative solution that does not require additional computation to generate synthetic samples. For example, for image classification tasks we may use large open datasets such as ImageNet~\cite{dengImageNetLargescaleHierarchical2009} as a substitute for the original data. While the images may represent different classes, a large dataset of diverse images may be sufficient to distill knowledge from the expert models. This technique is also more efficient since it does not require a separate data generation phase. However, it is possible to use it only if a large open dataset is available, which may be true for image classification problems with natural images but more difficult in other domains, such as the medical domain, where data is scarcer. Since data comes from a different domain than the original one, we do not have a target class corresponding to the original domain. Therefore, we set $\vy^{syn} = \argmax \fexpert_i(\vx^{syn})$ for each sample $\vx^{syn}$ in the auxiliary dataset.

\subsection{Natural Image Priors}
Synthetic data generation methods tend to generate images with unrealistic artifacts. The end-to-end optimization over the raw pixels generates images with high output probabilities in the target classes that do not resemble the original images. Natural image priors are regularization terms that encourage natural looking images.

\vspace{-12pt}
\paragraph{Augmentations} All the synthetic images are augmented with common image augmentations, both during the generation and during the ex-model distillation. Depending on the dataset, we use small displacements, rotations, and horizontal flip.

\vspace{-12pt}
\paragraph{$L^2$ norm penalization} We penalize the $L^2$ norm of each image $\loss_{norm}(\vx^{syn}  ) = \norm{\vx^{syn}}^2_2$. This regularization term is used to penalize high activations.

\vspace{-12pt}
\paragraph{Blur} In natural images, neighboring pixels are similar. To encourage this property, we penalize the term $\loss_{blur}(\vx^{syn}) = \norm{\vx^{syn} - blur(\vx^{syn}) }^2_2$, where $blur(\vx^{syn})$ is the result of applying a gaussian blur with a $3 \times 3$ kernel to the raw image $\vx^{syn}$.

\vspace{-12pt}
\paragraph{Matching of batch normalization statistics} Batch normalization layers provide useful information about the activations' statistic of the original images~\cite{yinDreamingDistillDatafree2020}. Ideally, synthetic images should have the same statistics. Given a model with $k$ batch normalization layers with mean and variance $\mu_i, \sigma_i$, and minibatch statistic for the synthetic images $\mu^{syn}_i, \sigma^{syn}_i$, we penalize the term $\loss_{bns} = \sum_{i=0}^{k} (\norm{\mu_i - \mu^{syn}_i}^2_2 + \norm{\sigma_i - \sigma^{syn}_i}^2_2$).

%%%%%%%%%%%%%%%%%%%%%%%%%%%%%%%%%%%%%%%%%%%%%%%%%%%%%%%%%%%%%%%%%%%%%
\section{Experiments}
\label{sec:experiments}
%%%%%%%%%%%%%%%%%%%%%%%%%%%%%%%%%%%%%%%%%%%%%%%%%%%%%%%%%%%%%%%%%%%%%
The objective of the experimental evaluation is twofold: first, we evaluate ex-model scenarios under different conditions by proposing novel \exml benchmarks with varying levels of complexity. Second, we assess the performance of different ex-model distillation strategies by comparing different sources of synthetic data as defined in Section \ref{sec:datagen}, along with a set of baselines.

For each scenario, we trained a separate expert model for each learning experience using the original data. For each configuration, we repeated the training phase 5 times to obtain 5 independent streams of experts that we used to compute the mean and standard deviation. For simplicity, we use the same architecture during the ex-model continual learning phase. 
All the experiments are implemented using Avalanche~\cite{lomonacoAvalancheEndtoEndLibrary2021}. Source code for the proposed strategies, along with the experiments configuration, code to reproduce the experiments, and the pretrained experts is available online\footnote{\url{https://github.com/AntonioCarta/ex_model_cl}}. Please refer to the repository and the additional material for extended details about the hyperparameters of the experiments.
\begin{table}[]
    \small
    \centering
    \caption{Summary of datasets and scenarios.}\label{tbl:data}
    \scalebox{0.93}{    
    \begin{tabular}{lcccl}
    \toprule
    scenario        & \shortstack{stream \\ length}    & \shortstack{total \\ classes} & \shortstack{classes \\ per step} &  model \\ \midrule
    MNIST-NC            & 5         & 10      & 2            & LeNet         \\
    CIFAR10-NC          & 10        & 10      & 2            & ResNet18      \\
    CIFAR10-MT          & 10        & 10      & 2            & ResNet18      \\
    CORe50-NC           & 9         & 50      & 10/5         & MobileNet     \\
    CORe50-NI           & 9         & 50      & 50           & MobileNet     \\ \bottomrule
    \end{tabular}}
\end{table}
\paragraph{Datasets and continual learning scenarios}

\begin{table*}[t]
    \centering
    \caption{Stream accuracy computed on the test set for MNIST and CIFAR10 continual learning scenarios. Ensemble methods' results are not shown for joint scenarios because ensembling is not necessary when there is a single model.}
    \label{tab:res_cifar}
    \begin{tabular}{lcrrrrr}
        \toprule
                            & \multirow{2}{*}{\shortstack{Ex-model \\ scenario}} & \multicolumn{2}{c}{MNIST} & \multicolumn{3}{c}{CIFAR10} \\
                    &        & Joint               & NC               & Joint               & NC               & MT \\ \midrule
Oracle              & \xmark & \mustd{93.71}{0.28} & \mustd{99.42}{0.19} & \mustd{87.37}{1.11} & \mustd{96.58}{0.86} & \mustd{96.58}{0.86} \\ 
Ensemble Avg.       & \xmark & \emptycell          & \mustd{33.40}{4.74} & \emptycell          & \mustd{51.85}{2.37} & \emptycell \\
Min Entropy         & \xmark & \emptycell          & \mustd{39.41}{5.27} & \emptycell          & \mustd{52.03}{2.67} & \emptycell \\ \midrule
Param. Avg.         & \cmark & \emptycell          & \mustd{20.11}{0.97} & \emptycell          & \mustd{10.00}{0.00} & \mustd{51.85}{2.37} \\ \midrule
% Replay ED           & \xmark & \mustd{96.11}{0.18} & \mustd{76.46}{3.87} & \mustd{80.29}{1.09} & \mustd{76.66}{1.06} & \mustd{}{} \\ \midrule
% Gaussian ED         & \cmark & \mustd{09.69}{6.46} & \mustd{}{}          & \mustd{}{}          & \mustd{}{} & \mustd{}{} \\
Model Inversion ED  & \cmark & \mustd{93.09}{1.43} & \mustd{43.23}{3.00} & \mustd{64.55}{3.25} & \mustd{17.40}{3.96} & \mustd{61.71}{7.52} \\
Data Impression ED  & \cmark & \mustd{92.12}{0.88} & \mustd{36.05}{6.74} & \mustd{52.64}{5.82} & \mustd{24.70}{6.85} & \mustd{61.15}{3.92} \\
Aux. Data ED        & \cmark & \mustd{89.35}{0.18} & \mustd{35.48}{6.35} & \mustd{76.94}{2.68} & \mustd{41.35}{5.83} & \mustd{60.72}{3.70} \\
        \bottomrule
        
    \end{tabular}
\end{table*}

In each experiment, the stream of experts is trained on popular CL benchmarks. Most benchmarks come from the class-incremental literature~\cite{masanaClassincrementalLearningSurvey2020}, where each experience provides data for New Classes (NC), never seen before. We also ran experiments on the New Instances (NI) scenario~\cite{pmlr-v78-lomonaco17a}, where each experience has the same classes with different instances (e.g. different backgrounds). Therefore, in the NI scenario expert models are trained on the same set of classes.
Additionally, we show the results for joint training, i.e. the offline training where the data is seen all at once. In this setting, we do not have a stream of pretrained models. We used this scenario to evaluate the performance of the data extraction methods in the absence of continual learning.

We evaluated the proposed strategies on MNIST~\cite{lecunMNISTDatabaseHandwritten1998} with a stream of LeNet~\cite{krizhevskyImagenetClassificationDeep2012} models, using the Split MNIST (NC) scenario, with 5 experiences and 2 classes for each experience. We also provide the results for joint training to evaluate the degradation in performance from a simple data-free knowledge distillation to a more challenging Ex-Model CL scenario. For CIFAR10 ~\cite{krizhevskyLearningMultipleLayers}, we used a ResNet18~\cite{heDeepResidualLearning2016} and we evaluated both the popular joint training scenario and the Split-CIFAR10 (NC) scenario~\cite{masanaClassincrementalLearningSurvey2020}. The joint scenario uses all 10 classes at once, while the class incremental scenario uses 2 classes per experience. Furthermore, we evaluate Split-CIFAR10 in a multitask setting with a multi-head classifier (CIFAR10-MT). 

Finally, we used CORe50~\cite{pmlr-v78-lomonaco17a}, a dataset specifically designed for continual learning. In the joint scenario we used all the 50 classes of the dataset, while in the class-incremental (NC) scenario we used 10 classes for the first experience and 5 for the subsequent ones. We also experimented with CORe50 in the NIC scenario. For all the configurations, we used a MobileNet~\cite{howardMobileNetsEfficientConvolutional2017} pretrained on ImageNet.

Ex-model strategies are evaluated in the single incremental task (SIT) setting with a single head: this means that the model does not have a task label to distinguish between the different experiences. This is the most challenging setting. A summary of the configuration for each scenario is shown in Table \ref{tbl:data}.

\paragraph{Ex-model strategies} 

We evaluated three Ex-model Distillation (ED) strategies: Model Inversion ED, Data Impression ED, Aux. Data ED. Each one uses a different source for synthetic data. All data extraction strategies use a memory buffer with a fixed size (5000 samples) to maintain the data extracted from the previous experiences while keeping the memory occupation reasonable for continual learning on a large stream of experts. We use Fashion MNIST~\cite{xiaoFashionMNISTNovelImage2017} as auxiliary data for MNIST scenarios, and ImageNet for CIFAR10 and CORe50.
Additionally, we show the results of four baselines. Except Param. Avg., all the baselines do not satisfy the constraints of the ex-model scenario since they store the full stream of models or require the original data. 
\begin{description}
  \item[Oracle] This is an ensemble of the stream of experts. The ensemble uses a task label to determine the correct model to use. Notice that this ensemble achieves higher results than offline training since the task label makes the classification easier by excluding all the classes corresponding to different tasks.
  \item[Ensemble Avg.] This is an ensemble of experts which computes the output as the average of the experts' outputs.
  \item[Min. Entropy] This is a strategy that computes the output for each expert in the stream and uses as final output only the prediction with the minimum entropy (i.e. the one with less uncertainty). Given the output probabilities $\vct{p}^j$ computed by the $j$-th expert, the ensemble selects the outputs from expert $j^*$, where $j^* = \argmin_j - \sum_i p^j_i \log p^j_i$. This ensemble computes all the output in parallel, similarly to the Oracle ensemble. However, the entropy is used to select the appropriate expert instead of the task label.
  \item[Param. Avg.] This ensemble is a single model obtained by averaging the expert's parameters. This is the only baseline which respects the ex-model scenario constraints since it keeps a single model and it does not use the original data. Unlike ED, this strategy assumes that the experts' architectures are all equal.
  \item[Replay ED] This is a strategy where the ex-model distillation is applied using the original data. Notice that this is different from a simple rehearsal strategy since it uses the loss of Eq.~\ref{eq:ed_loss} instead of the crossentropy.
\end{description}

\begin{table*}[t]
    \centering
    \caption{Stream accuracy computed on the test set for CORe50 continual learning scenarios. Ensemble methods' results are not shown for joint scenarios because ensembling is not necessary when there is a single model.}
    \label{tab:res_core50}
    \begin{tabular}{lcrrrrrr}
        \toprule
                            & \multirow{2}{*}{\shortstack{Ex-model \\ scenario}} & \multicolumn{3}{c}{CORe50} \\
                            &        & Joint & NC & NI \\ \midrule
        Oracle              & \xmark & \mustd{85.73}{0.29} & \mustd{96.04}{1.08} & \emptycell \\ 
        Ensemble Avg.       & \xmark & \emptycell          & \mustd{26.30}{1.38} & \mustd{69.92}{0.70} \\
        Min. Entropy        & \xmark & \emptycell          & \mustd{42.41}{0.96} & \mustd{61.36}{1.86} \\ \midrule
        Param. Avg.         & \cmark & \emptycell          & \mustd{2.00}{0.00}          & \mustd{2.00}{0.00} \\ \midrule
        % Replay ED           & \xmark & \mustd{87.44}{0.28} & \mustd{80.75}{0.65} & \mustd{83.78}{0.56} \\ \midrule
        % Gaussian ED         & \cmark & \mustd{}{}          & \mustd{}{}          & \mustd{}{} \\
        Model Inversion ED  & \cmark & \mustd{50.06}{2.76} & \mustd{33.1}{1.93}  & \mustd{44.38}{4.93} \\
        Data Impression ED  & \cmark & \mustd{52.91}{2.09} & \mustd{17.57}{3.57} & \mustd{43.26}{2.36} \\
        Aux. Data ED        & \cmark & \mustd{81.82}{0.29} & \mustd{34.87}{1.16} & \mustd{44.51}{2.91} \\
        \bottomrule
    \end{tabular}
\end{table*}

Ensemble baselines show the performance that can be obtained by combining the expert models without any training. The memory requirements of an ensemble grows linearly in the number of models, making these strategies not admissible for an ex-model scenario. Instead, Replay ED shows the performance of the ex-model distillation in the ideal setting where we have access to the original data.

\subsection{Results}

% data generation under different scenarios
\begin{figure*}
  \begin{subfigure}{.19\textwidth}
    \centering
    \includegraphics[width=\linewidth]{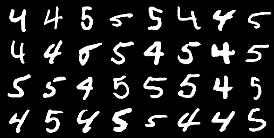}  
    \caption{Original data. \\\hspace{\textwidth}}\label{fig:sub-1}
  \end{subfigure}
  \begin{subfigure}{.19\textwidth}
    \centering
    \includegraphics[width=\linewidth]{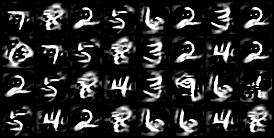}  
    \caption{Joint MNIST, \\ Model Inversion}\label{fig:sub-2}
  \end{subfigure}
  \begin{subfigure}{.19\textwidth}
    \centering
    \includegraphics[width=\linewidth]{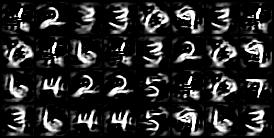}  
    \caption{Joint MNIST, \\ Data Impression}\label{fig:sub-3}
  \end{subfigure}
  \begin{subfigure}{.19\textwidth}
    \centering
    \includegraphics[width=\linewidth]{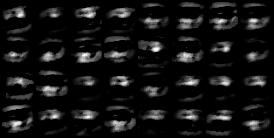}  
    \caption{Split MNIST, \\ Model Inversion}\label{fig:sub-4}
  \end{subfigure}
  \begin{subfigure}{.19\textwidth}
    \centering
    \includegraphics[width=\linewidth]{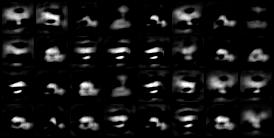}  
    \caption{Split MNIST, \\ Data Impression}\label{fig:sub-5}
  \end{subfigure}
  \caption{Original data and generated samples for Joint MNIST and Split MNIST.}\label{fig:gensamples}
\end{figure*}

% results on MNIST/CIFAR10
Table~\ref{tab:res_cifar} shows the stream accuracy on the test set for MNIST and CIFAR10 scenarios. When we have access to the entire stream of expert models and task labels (Oracle) we obtain the upper bound performance. Notice that the accuracy of this strategy in the class incremental scenarios is even higher than the joint scenarios since the task labels provide additional information that restricts the number of possible classes for a given sample. Ensemble methods have a large drop in performance in the class incremental scenarios. Notice that we do not consider them to be proper ex-model strategies since they keep the entire stream of models. The only proper ex-model ensemble strategy is the Param. Avg., which obtains a random performance on CIFAR10. This is due to the large number of models that need to be averaged together, each one trained on different data. Ex-model distillation strategies have a very high performance in joint scenarios, showing that synthetic and auxiliary data is sufficient to perform the knowledge distillation. However, there is a large performance drop in the class incremental scenarios, except for CIFAR10-MT, which is multi-task and therefore easier to learn.
Table \ref{tab:res_core50} shows the stream accuracy on the test set for CORe50 scenarios. CORe50 is more challenging than MNIST and CIFAR10 due to the higher resolution images (we use 128x128 images in our experiments). However, since we used a pretrained MobileNet, all the expert models start from a common initialization with a rich feature extractor. The initialization helps to learn and mitigate the interference between the experts in the NC and NI scenarios. Notice that in the NI scenario it is not possible to evaluate the Oracle baseline since CORe50-NI provides a single test set that cannot be split, unlike the NC scenario where we split by classes.

% results on CORE50

\paragraph{Buffer samples} Figure \ref{fig:gensamples} shows the samples generated by model inversion and data impression on joint MNIST and Split MNIST. In both settings, the images have been trained until the desired class was predicted with probability $> 0.99$. Both methods generate visually plausible digits in the joint scenario, wich resemble multiple digits superimposed over each other. Instead, in the class incremental scenarios, despite the high confidence of the model, the images are far from being realistic. This result may partially explain the different results of joint and continual learning scenarios. In joint scenarios, the single expert must be able to distinguish between all the possible classes. Instead, in continual scenarios, the experts will overfit their small subset of data, and will incorrectly classify out-of-domain classes with high confidence. As a result, the generated images will not be realistic because the model did not learn to extract the features that would help to classify the images in the joint domain. 

For more samples please check the additional material. While the difference between the joint and NC scenarios is less striking in some CIFAR10 and CORe50 configurations, the general conclusions are the same and it can be easily noticed that images generated in the NC scenario are qualitatively worse.

\begin{figure}[t]
    \centering
    \includegraphics[width=.5\textwidth]{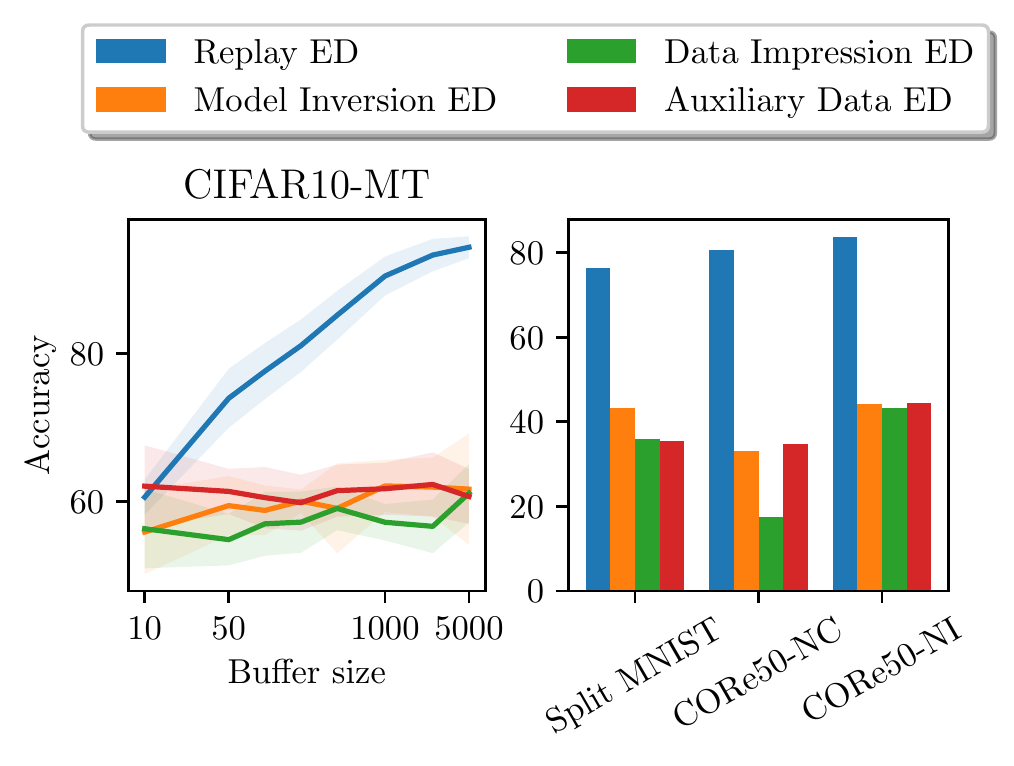}
    \caption{Test stream accuracy on CIFAR10-MT for increasing buffer sizes (left). Results are shown on a semi-logarithmic scale. Comparison between different strategies, including ex-model distillation using a small buffer of the original data (Replay ED, on the right).}
    \label{fig:ablation}
\end{figure}

\paragraph{Buffer size} Figure \ref{fig:ablation} shows the test stream accuracy on CIFAR10-MT for increasing buffer sizes. The minimum size is $10$, corresponding to a single sample per class. Notice that the blue line showing ex-model distillation using a subset of the original data increases with larger buffers. Instead, other methods show negligible differences between the minimum and maximum buffer size. This results hints that the major limitation of current data extraction techniques is the scaling to larger buffers. There may be several reasons for the lack of scaling in accuracy. For example, the diversity between generated images of the same class may be insufficient, which renders large buffers useless (see the additional material for some samples). 

\paragraph{Buffer strategy}
Figure \ref{fig:ablation} shows the performance of ex-model distillation techniques against ex-model distillation on the real data (Replay ED). We notice that there is a large gap between Replay ED and proper ex-model distillation strategies.

Overall, we did not find a large performance difference between Model Inversion and Data Impression. Instead, we see techniques based on data generation perform better on MNIST, while auxiliary data is better on some CIFAR10 and CORe50 scenarios. We argue that this is a consequence of the similarity between the original data and the auxiliary data. MNIST and Fashion MNIST are very different, except for the fact that they both use greyscale images, since the domains are completely separate. Instead, CIFAR10 and CORe50 are much closer to ImageNet since they both contains natural images, albeit with different resolutions and classes. Furthermore, it appears synthetic data techniques perform much better on multitask scenarios.

\section{Related Works}
% continual learning
% scenari simili: data-free CIL
The objective of continual learning is to continuously adapt the model without forgetting the previous knowledge \cite{lesortContinualLearningRobotics2019}. Recently, there has been an increasing interest towards data-free settings, where the previous data is not available. Under this constraint, the most popular scenario is the data-free class-incremental learning (DF-CIL), where new classes appear over time. Notice that, despite the apparent similarity, DF-CIL is a very different scenario from ex-model. First, in DF-CIL the current data is available, making it possible to exploit a subset of the real data. Furthermore, in DF-CIL the model is trained sequentially, while in ex-model scenarios the experts are trained independently.

% knowledge distillation
% double distillation
Learning without forgetting (LwF) \cite{liLearningForgetting2017} is a continual learning strategy that mitigates catastrophic forgetting via knowledge distillation of the old model's logits computed on the new data. More recently, several proposals have adapted knowledge distillation to DF-CIL scenarios. \cite{zhangClassincrementalLearningDeep2020} exploits publicly available training data. \cite{smithAlwaysBeDreaming2021, choiDualTeacherClassIncrementalLearning2021} train generative models to extract synthetic samples.

% image prior, gen img
Data-free knowledge distillation methods apply several techniques to generate representative and diverse samples. Image priors help to guide the optimization process towards natural looking images during the model inversion\cite{dosovitskiyInvertingVisualRepresentations2016}. \cite{mahendranUnderstandingDeepImage2014} proposes the use of norm penalization to penalize high activations and total variation to penalize differences between small shifts. \cite{yinDreamingDistillDatafree2020} proposes to match batch normalization statistics between the real and generated data. 
\cite{nayakZeroShotKnowledgeDistillation2019} generates targets logits according to a Dirichlet distribution instead of hard targets in order to capture inter-class similarities, while \cite{yooKnowledgeExtractionNo} trains generative networks instead of the samples directly.
Notice that realistic images are not strictly necessary to perform knowledge distillation. \cite{beyerKnowledgeDistillationGood2021} shows that knowedge distillation can be modeled as function matching. They show that aggressive augmentations combined with long training regimes help the knowledge distillation.

\section{Discussion and Conclusion}
\label{sec:discussions}

In this paper we introduced Ex-Model Continual Learning, a novel scenario to continuously train a model from a stream of pretrained experts, without assuming any access to training data. We proposed a family of continual learning strategies, called Ex-Model Distillation, able to transfer knowledge from the experts to the Ex-Model, trained continuously. We validated the ability of three ED strategies to learn in our novel scenario against three different continual learning benchmarks.
% Future works may improve Ex-Model Distillation by designing either new distillation approaches or new algorithms for the generation of a more diverse buffer of synthetic data. \\
\exml exploits the growing number of pretrained models currently available for many different applications (object detection, language modelling...), without making any assumptions on the model architecture or the training modalities. \exml would benefit from an organized categorization of the existing pretrained models and of the type of knowledge they acquired during training. Such \emph{neural skills catalogue} would make it easier to decide, when possible, which expert to select in order to best incorporate the required knowledge. \\
\exml is related to modern distributed learning paradigms, where different models are trained independently and their knowledge is then aggregated into a centralized architecture. Unlike federated learning, where each agent is constantly communicating with a centralized server, in \exml each agent is independent and the communication between agents is limited. Moreover, as privacy-aware settings are gaining relevance within the machine learning community, the need to learn in data-free environment will become mandatory for many applications. Medical environments, for example, are often subjected to strong privacy constraints, where it might not be possible to transfer data collected from patients to other devices. Ultimately, \exml constitutes a novel paradigm which does not supersede the available continual learning scenarios, but instead it stands as promising alternative to deliver continual learning capabilities to otherwise inaccessible real-world environments.

%Ex-model CL opens many avenues for future research. For example, it is possible to devise strategies for specific applications. In federated and privacy-aware settings, we may assume that the strategy controls the training of each expert, and therefore we may reduce the interference between the experts. Instead, in scenarios where we want to integrate a large pretrained model, we may not have access to its data, but we may have access to the original data for all the other experiences. Finally, it may be possible to improve the ex-model distillation by extracting more natural looking images.

%%%%%%%%% REFERENCES
{\small
\bibliographystyle{ieee_fullname}
\bibliography{egbib,biblio}

\begin{thebibliography}{10}\itemsep=-1pt

\bibitem{abadiDeepLearningDifferential2016}
Martin Abadi, Andy Chu, Ian Goodfellow, H.~Brendan McMahan, Ilya Mironov, Kunal
  Talwar, and Li Zhang.
\newblock Deep {{Learning}} with {{Differential Privacy}}.
\newblock In {\em Proceedings of the 2016 {{ACM SIGSAC Conference}} on
  {{Computer}} and {{Communications Security}}}, {{CCS}} '16, pages 308--318,
  {New York, NY, USA}, Oct. 2016. {Association for Computing Machinery}.

\bibitem{beyerKnowledgeDistillationGood2021}
Lucas Beyer, Xiaohua Zhai, Am{\'e}lie Royer, Larisa Markeeva, Rohan Anil, and
  Alexander Kolesnikov.
\newblock Knowledge distillation: A good teacher is patient and consistent.
\newblock {\em arXiv:2106.05237 [cs]}, June 2021.

\bibitem{choiDualTeacherClassIncrementalLearning2021}
Yoojin Choi, Mostafa {El-Khamy}, and Jungwon Lee.
\newblock Dual-{{Teacher Class}}-{{Incremental Learning With Data}}-{{Free
  Generative Replay}}.
\newblock {\em arXiv:2106.09835 [cs]}, June 2021.

\bibitem{dengImageNetLargescaleHierarchical2009}
Jia Deng, Wei Dong, Richard Socher, Li-Jia Li, Kai Li, and Li {Fei-Fei}.
\newblock {{ImageNet}}: A large-scale hierarchical image database.
\newblock In {\em 2009 {{IEEE Conference}} on {{Computer Vision}} and {{Pattern
  Recognition}}}, pages 248--255, June 2009.

\bibitem{dosovitskiyInvertingVisualRepresentations2016}
Alexey Dosovitskiy and Thomas Brox.
\newblock Inverting {{Visual Representations}} with {{Convolutional Networks}}.
\newblock {\em arXiv:1506.02753 [cs]}, Apr. 2016.

\bibitem{hayesLifelongMachineLearning2020a}
Tyler~L. Hayes and Christopher Kanan.
\newblock Lifelong {{Machine Learning}} with {{Deep Streaming Linear
  Discriminant Analysis}}.
\newblock In {\em 2020 {{IEEE}}/{{CVF Conference}} on {{Computer Vision}} and
  {{Pattern Recognition Workshops}} ({{CVPRW}})}, pages 887--896, {Seattle, WA,
  USA}, June 2020. {IEEE}.

\bibitem{hayesReplayDeepLearning2021}
Tyler~L. Hayes, Giri~P. Krishnan, Maxim Bazhenov, Hava~T. Siegelmann,
  Terrence~J. Sejnowski, and Christopher Kanan.
\newblock Replay in {{Deep Learning}}: Current {{Approaches}} and {{Missing
  Biological Elements}}.
\newblock {\em arXiv:2104.04132 [cs, q-bio]}, Apr. 2021.

\bibitem{heDeepResidualLearning2016}
Kaiming He, Xiangyu Zhang, Shaoqing Ren, and Jian Sun.
\newblock Deep residual learning for image recognition.
\newblock In {\em Proceedings of the {{IEEE Conference}} on {{Computer Vision}}
  and {{Pattern Recognition}}}, pages 770--778, 2016.

\bibitem{hendryxFederatedReconnaissanceEfficient2021}
Sean~M. Hendryx, Dharma~Raj KC, Bradley Walls, and Clayton~T. Morrison.
\newblock Federated {{Reconnaissance}}: Efficient, {{Distributed}},
  {{Class}}-{{Incremental Learning}}.
\newblock {\em arXiv:2109.00150 [cs]}, Aug. 2021.

\bibitem{hintonDistillingKnowledgeNeural2015}
Geoffrey Hinton, Oriol Vinyals, and Jeff Dean.
\newblock Distilling the {{Knowledge}} in a {{Neural Network}}.
\newblock pages 1--9, 2015.

\bibitem{howardMobileNetsEfficientConvolutional2017}
Andrew~G. Howard, Menglong Zhu, Bo Chen, Dmitry Kalenichenko, Weijun Wang,
  Tobias Weyand, Marco Andreetto, and Hartwig Adam.
\newblock {{MobileNets}}: Efficient {{Convolutional Neural Networks}} for
  {{Mobile Vision Applications}}.
\newblock {\em arXiv:1704.04861 [cs]}, Apr. 2017.

\bibitem{kemkerFearNetBrainInspiredModel2018}
Ronald Kemker and Christopher Kanan.
\newblock {{FearNet}}: Brain-{{Inspired Model}} for {{Incremental Learning}}.
\newblock In {\em International {{Conference}} on {{Learning
  Representations}}}, Feb. 2018.

\bibitem{krizhevskyLearningMultipleLayers}
Alex Krizhevsky.
\newblock Learning {{Multiple Layers}} of {{Features}} from {{Tiny Images}}.
\newblock page~60.

\bibitem{krizhevskyImagenetClassificationDeep2012}
Alex Krizhevsky, Ilya Sutskever, and Geoffrey~E Hinton.
\newblock Imagenet classification with deep convolutional neural networks.
\newblock In {\em Advances in Neural Information Processing Systems}, pages
  1097--1105, 2012.

\bibitem{lecunMNISTDatabaseHandwritten1998}
Yann LeCun.
\newblock The {{MNIST}} database of handwritten digits.
\newblock {\em http://yann. lecun. com/exdb/mnist/}, 1998.

\bibitem{lesortContinualLearningRobotics2019}
Timoth{\'e}e Lesort, Vincenzo Lomonaco, Andrei Stoian, Davide Maltoni, David
  Filliat, and Natalia {D{\'i}az-Rodr{\'i}guez}.
\newblock Continual {{Learning}} for {{Robotics}}: Definition, {{Framework}},
  {{Learning Strategies}}, {{Opportunities}} and {{Challenges}}.
\newblock {\em arXiv:1907.00182 [cs]}, Nov. 2019.

\bibitem{liLearningForgetting2017}
Zhizhong Li and Derek Hoiem.
\newblock Learning without {{Forgetting}}.
\newblock {\em arXiv:1606.09282 [cs, stat]}, Feb. 2017.

\bibitem{pmlr-v78-lomonaco17a}
Vincenzo Lomonaco and Davide Maltoni.
\newblock {{CORe50}}: A new dataset and benchmark for continuous object
  recognition.
\newblock In Sergey Levine, Vincent Vanhoucke, and Ken Goldberg, editors, {\em
  Proceedings of the 1st Annual Conference on Robot Learning}, volume~78 of
  {\em Proceedings of Machine Learning Research}, pages 17--26. {PMLR}, Nov.
  2017.

\bibitem{lomonacoAvalancheEndtoEndLibrary2021}
Vincenzo Lomonaco, Lorenzo Pellegrini, Andrea Cossu, Antonio Carta, Gabriele
  Graffieti, Tyler~L. Hayes, Matthias De~Lange, Marc Masana, Jary Pomponi,
  Gido~M. {van de Ven}, Martin Mundt, Qi She, Keiland Cooper, Jeremy Forest,
  Eden Belouadah, Simone Calderara, German~I. Parisi, Fabio Cuzzolin,
  Andreas~S. Tolias, Simone Scardapane, Luca Antiga, Subutai Ahmad, Adrian
  Popescu, Christopher Kanan, Joost {van de Weijer}, Tinne Tuytelaars, Davide
  Bacciu, and Davide Maltoni.
\newblock Avalanche: An {{End}}-to-{{End Library}} for {{Continual Learning}}.
\newblock In {\em Proceedings of the {{IEEE}}/{{CVF Conference}} on {{Computer
  Vision}} and {{Pattern Recognition}}}, pages 3600--3610, 2021.

\bibitem{mahendranUnderstandingDeepImage2014}
Aravindh Mahendran and Andrea Vedaldi.
\newblock Understanding {{Deep Image Representations}} by {{Inverting Them}}.
\newblock {\em arXiv:1412.0035 [cs]}, Nov. 2014.

\bibitem{maltoniContinuousLearningSingleincrementaltask2019}
Davide Maltoni and Vincenzo Lomonaco.
\newblock Continuous learning in single-incremental-task scenarios.
\newblock {\em Neural Networks}, 116:56--73, Aug. 2019.

\bibitem{masanaClassincrementalLearningSurvey2020}
Marc Masana, Xialei Liu, Bartlomiej Twardowski, Mikel Menta, Andrew~D.
  Bagdanov, and Joost {van de Weijer}.
\newblock Class-incremental learning: Survey and performance evaluation.
\newblock {\em arXiv:2010.15277 [cs]}, Oct. 2020.

\bibitem{nayakZeroShotKnowledgeDistillation2019}
Gaurav~Kumar Nayak, Konda~Reddy Mopuri, Vaisakh Shaj, R.~Venkatesh Babu, and
  Anirban Chakraborty.
\newblock Zero-{{Shot Knowledge Distillation}} in {{Deep Networks}}.
\newblock {\em arXiv:1905.08114 [cs, stat]}, May 2019.

\bibitem{parisiContinualLifelongLearning2019}
German~I. Parisi, Ronald Kemker, Jose~L. Part, Christopher Kanan, and Stefan
  Wermter.
\newblock Continual lifelong learning with neural networks: A review.
\newblock {\em Neural Networks}, 113:54--71, Feb. 2019.

\bibitem{ringCHILDFirstStep1997}
Mark~B. Ring.
\newblock {{CHILD}}: A {{First Step Towards Continual Learning}}.
\newblock {\em Machine Learning}, 28(1):77--104, July 1997.

\bibitem{smithAlwaysBeDreaming2021}
James Smith, Yen-Chang Hsu, Jonathan Balloch, Yilin Shen, Hongxia Jin, and
  Zsolt Kira.
\newblock Always {{Be Dreaming}}: A {{New Approach}} for {{Data}}-{{Free
  Class}}-{{Incremental Learning}}.
\newblock {\em arXiv:2106.09701 [cs]}, June 2021.

\bibitem{usmanovaDistillationbasedApproachIntegrating2021}
Anastasiia Usmanova, Fran{\c c}ois Portet, Philippe Lalanda, and German Vega.
\newblock A distillation-based approach integrating continual learning and
  federated learning for pervasive services.
\newblock {\em arXiv:2109.04197 [cs]}, Sept. 2021.

\bibitem{xiaoFashionMNISTNovelImage2017}
Han Xiao, Kashif Rasul, and Roland Vollgraf.
\newblock Fashion-{{MNIST}}: A {{Novel Image Dataset}} for {{Benchmarking
  Machine Learning Algorithms}}.
\newblock {\em arXiv:1708.07747 [cs, stat]}, Sept. 2017.

\bibitem{yinDreamingDistillDatafree2020}
Hongxu Yin, Pavlo Molchanov, Zhizhong Li, Jose~M. Alvarez, Arun Mallya, Derek
  Hoiem, Niraj~K. Jha, and Jan Kautz.
\newblock Dreaming to {{Distill}}: Data-free {{Knowledge Transfer}} via
  {{DeepInversion}}.
\newblock {\em arXiv:1912.08795 [cs, stat]}, June 2020.

\bibitem{yooKnowledgeExtractionNo}
Jaemin Yoo, Minyong Cho, Taebum Kim, and U Kang.
\newblock Knowledge {{Extraction}} with {{No Observable Data}}.
\newblock page~10.

\bibitem{yoonFederatedContinualLearning2021}
Jaehong Yoon, Wonyong Jeong, Giwoong Lee, Eunho Yang, and Sung~Ju Hwang.
\newblock Federated {{Continual Learning}} with {{Weighted Inter}}-client
  {{Transfer}}.
\newblock {\em arXiv:2003.03196 [cs, stat]}, June 2021.

\bibitem{zhangClassincrementalLearningDeep2020}
Junting Zhang, Jie Zhang, Shalini Ghosh, Dawei Li, Serafettin Tasci, Larry
  Heck, Heming Zhang, and C.-C.~Jay Kuo.
\newblock Class-incremental {{Learning}} via {{Deep Model Consolidation}}.
\newblock {\em arXiv:1903.07864 [cs]}, Jan. 2020.

\end{thebibliography}
}

% do we have an appendix @ CVPR?
\appendix 
\section{Benchmarks}
% original data and experts hyperparameters
% TODO: ex-model hyperparameters
The streams of pretrained expert models to reproduce our results can be downloaded on the source code repository. The training loop of each expert uses an SGD optimizer with learning rate 0.1, momentum 0.9, and weight decay $5e-4$, batch size $32$, and $10$ training epochs. 

\subsection{MNIST}
\begin{description}
    \item[Joint MNIST] A single experience ($50000$ samples, $10$ classes).
    \item[Split MNIST] Class-incremental scenario. A stream of $5$ experiences, each with $2$ classes and $10000$ samples.
\end{description}

Hyperparameters used to train the experts on MNIST scenarios:

\begin{center}
\begin{tabular}{lrr}
    \toprule
    & \multicolumn{2}{c}{MNIST} \\
                & Joint & NC \\ \midrule        
    epochs        & 100   & 20   \\
    learning rate & 0.01  & 0.01 \\ \bottomrule   
\end{tabular}
\end{center}

\subsection{CIFAR10}
\begin{description}
    \item[Joint CIFAR10] A single experience ($50000$ samples, $10$ classes).
    \item[Split CIFAR10] Class-incremental scenario. A stream of $5$ experiences, each with $2$ classes and $10000$ samples.
    \item[MT CIFAR10] A stream of $5$ experiences, each with $2$ classes and $10000$ samples. This scenario provides a task label.
\end{description}

Hyperparameters used to train the experts on Joint CIFAR10, Split CIFAR10, and MT CIFAR10:

\begin{center}
    \begin{tabular}{lr}
    \toprule
    \textbf{Hyperparameter} & \textbf{Value} \\ \midrule
    epochs & 100 \\
    learning rate & 0.01 \\ \bottomrule   
\end{tabular}
\end{center}

\subsection{CORe50}
\begin{description}
    \item[Joint CORe50] A single experience ($50$ classes).
    \item[NC CORe50] Class-incremental scenario. A stream of $9$ experiences, with $10$ classes for the first experience, and $5$ classes for the following ones.
    \item[NI CORe50] A stream of $8$ experiences with new instances from the same class (NI scenario).
\end{description}

Hyperparameters used to train the experts on CORe50 scenarios:

\begin{center}
\begin{tabular}{lrrr}
    \toprule
                  & \multicolumn{3}{c}{CORe50} \\ 
                  & Joint & NC    & NI     \\ \midrule
    epochs        & 40    & 4     & 10     \\
    batch size    & 128   & 128   & 128    \\
    learning rate & 0.001 & 0.001 & 0.001  \\ \bottomrule
\end{tabular}
\end{center}

\section{Model Selection and Hyperparameters}
Choosing the hyperparameters of the ex-model distillation is challenging due to the lack of the original data. We distinguish between hyperparameters of the data generation process, and hyperparameters of the distillation process. For the former, we can assume to have access to the original data of a single experience and use its patterns to optimize the data generation and fix the hyperparameters. For the distillation loss hyperparameters, we fix the hyperparameters without doing a model selection model selection by setting $\lambda_{CE} = 1$. The other hyperparameters are set by performing a small grid search over a single experience, similarly to the data generation hyperparameters. We use a fixed buffer size of $5000$ patterns for all the experiments. This hyperparameter is not meant to be tuned since larger buffer sizes will never achieve worse results. Instead, the buffer size can be adjusted to trade off the computational cost and memory usage against the average accuracy.

In the following subsections we list the hyperparameters of the best models. Notice that hyperparameters shared between different benchmarks are shown only once for each dataset at the beginning of the section.

Scenario hyperparameters:
\begin{description}
    \item[model architecture:] architecture of the experts.
\end{description}

Knowledge distillation hyperparameters:

\begin{center}
\begin{tabularx}{.45\textwidth}{lX}
    \toprule
    \textbf{Hyperparameter} & \textbf{Description} \\ \midrule
    lr                        & learning rate of the optimizer.      \\
    iterations per experience & number of iterations per experience. \\
    mini-batch size           & mini-batch size.                     \\
    KD temperature            & temperature of the KD loss.          \\ \bottomrule         
\end{tabularx}
\end{center}

Data generation hyperparameters:

\begin{center}
\begin{tabularx}{.45\textwidth}{lX}
    \toprule
    \textbf{Hyperparameter} & \textbf{Description} \\ \midrule
    buffer\_beta        & beta coefficient for the Dirichlet distribution (only for Data Impression). \\
    buffer\_blur        & regularization strength for the blur penalty.                               \\
    buffer\_bns         & regularization strength for the BNS penalty.                                \\
    buffer\_iter        & number of iterations.                                                       \\
    buffer\_lr          & learning rate.                                                              \\
    buffer\_temperature & temperature of the loss.                                                    \\
    buffer\_wd          & regularization strength for the weight decay.                               \\ \bottomrule
\end{tabularx}
\end{center}

\subsection{MNIST}
\begin{description}
    \item[model architecture:] LeNet 
\end{description}

Knowledge Distillation hyperparameters:

\begin{center}
\begin{tabular}{lr}
    \toprule
    \textbf{Hyperparameter} & \textbf{Value} \\ \midrule
    buffer\_lr          & 0.01  \\
    buffer\_iter        & 30000 \\
    buffer\_mb\_size    & 32    \\
    buffer\_temperature & 2.0   \\ \bottomrule
\end{tabular}
\end{center}

\subsubsection{Model Inversion ED}
Data generation hyperparameters:

\begin{center}
\begin{tabular}{lr}
    \toprule
    \textbf{Hyperparameter} & \textbf{Value} \\ \midrule
    lr             & 0.01  \\
    iterations     & 3000  \\
    weight blur    & 0.001 \\
    weight BNS     & 1.0   \\
    weight decay   & 0.001 \\
    KD temperature & 2.0   \\ \bottomrule
\end{tabular}
\end{center}

\subsubsection{Data Impression ED}
Data generation hyperparameters:

\begin{center}
\begin{tabular}{lr}
    \toprule
    \textbf{Hyperparameter} & \textbf{Value} \\ \midrule    
    buffer\_beta        & 10.0  \\
    buffer\_blur        & 0.001 \\
    buffer\_bns         & 1.0   \\
    buffer\_iter        & 3000  \\
    buffer\_lr          & 0.01  \\
    buffer\_temperature & 20.0  \\
    buffer\_wd          & 0.001 \\ \bottomrule
\end{tabular}
\end{center}

\subsection{CIFAR10}
\begin{description}
    \item[model architecture] ResNet18 
\end{description}

Knowledge Distillation hyperparameters:

\begin{center}
\begin{tabular}{lr}
    \toprule
    \textbf{Hyperparameter} & \textbf{Value} \\ \midrule   
    lr                        & 0.1   \\
    iterations per experience & 50000 \\
    mini-batch size           & 32    \\
    KD temperature            & 2.0   \\ \bottomrule
\end{tabular} 
\end{center}

\subsubsection{Model Inversion ED}
Data generation hyperparameters:

\begin{center}
\begin{tabular}{lr}
    \toprule
    \textbf{Hyperparameter} & \textbf{Value} \\ \midrule   
    buffer\_blur        & 0.01   \\
    buffer\_bns         & 0.0    \\
    buffer\_iter        & 500    \\
    buffer\_lr          & 0.1    \\
    buffer\_mb\_size    & 1024   \\
    buffer\_size        & 5000   \\
    buffer\_temperature & 2.0    \\
    buffer\_wd          & 0.0001 \\ \bottomrule
\end{tabular}
\end{center}

\subsubsection{Data Impression ED}
Data generation hyperparameters:

\begin{center}
\begin{tabular}{lr}
    \toprule
    \textbf{Hyperparameter} & \textbf{Value} \\ \midrule   
    buffer\_beta        & 10.0   \\
    buffer\_blur        & 0.01   \\
    buffer\_bns         & 0.0    \\
    buffer\_iter        & 500    \\
    buffer\_lr          & 0.1    \\
    buffer\_mb\_size    & 1024   \\
    buffer\_size        & 5000   \\
    buffer\_temperature & 2.0    \\
    buffer\_wd          & 0.0001 \\ \bottomrule
\end{tabular}
\end{center}

\subsection{CORe50}
\begin{description}
    \item[model architecture] MobileNet pretrained on ImageNet 
\end{description}

Knowledge Distillation hyperparameters:

\begin{center}
\begin{tabular}{lr}
    \toprule
    \textbf{Hyperparameter} & \textbf{Value} \\ \midrule  
    lr                        & 0.001 \\
    iterations per experience & 20000 \\
    mini-batch size           & 10    \\
    KD temperature            & 2.0   \\ \bottomrule
\end{tabular}
\end{center}

\subsubsection{Model Inversion ED}
Data generation hyperparameters:

\begin{center}
\begin{tabular}{lr}
    \toprule
    \textbf{Hyperparameter} & \textbf{Value} \\ \midrule  
    buffer\_beta        & 1.0    \\
    buffer\_blur        & 0.001  \\
    buffer\_bns         & 10.0   \\
    buffer\_iter        & 500    \\
    buffer\_lr          & 0.1    \\
    buffer\_mb\_size    & 128    \\
    buffer\_size        & 5000   \\
    buffer\_temperature & 2.0    \\
    buffer\_wd          & 0.0001 \\ \bottomrule
\end{tabular}
\end{center}

\subsubsection{Data Impression ED}
Data generation hyperparameters:

\begin{center}
\begin{tabular}{lr}
    \toprule
    \textbf{Hyperparameter} & \textbf{Value} \\ \midrule 
    buffer\_beta        & 0.1    \\
    buffer\_blur        & 0.001  \\
    buffer\_bns         & 10.0   \\
    buffer\_iter        & 500    \\
    buffer\_lr          & 0.1    \\
    buffer\_mb\_size    & 128    \\
    buffer\_size        & 5000   \\
    buffer\_temperature & 2.0    \\
    buffer\_wd          & 0.0001 \\ \bottomrule
\end{tabular}
\end{center}

\section{Synthetic Samples}
In this section we show side-by-side the original images and the images used by each Ex-Model Distillation strategy. Notice that the generated images are normalized, and therefore the colors may look unnatural in some cases. 

\clearpage
Split MNIST:
\centering
\begin{minipage}{.5\textwidth}
    \centering
    \includegraphics[width=.95\linewidth]{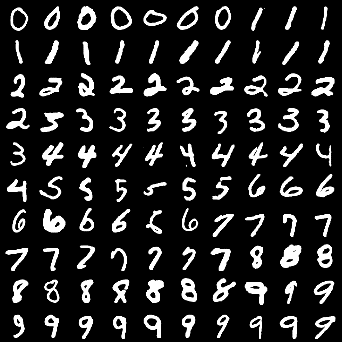}
    Original data.
\end{minipage}%
\begin{minipage}{.5\textwidth}
    \centering
    \includegraphics[width=.95\linewidth]{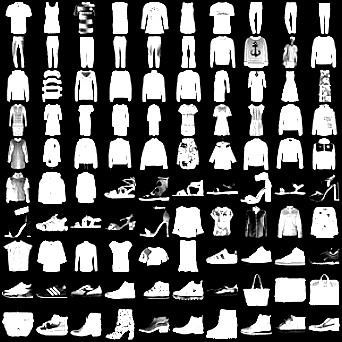}
    Auxiliary data (Fashion MNIST).
\end{minipage}
\begin{minipage}{.5\textwidth}
    \centering
    \includegraphics[width=.95\linewidth]{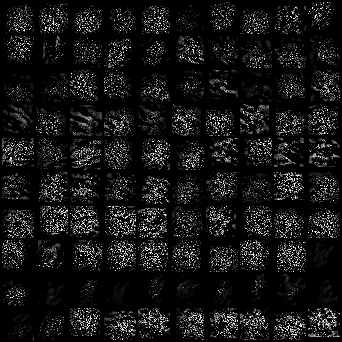}
    Model Inversion.
\end{minipage}%
\begin{minipage}{.5\textwidth}
    \centering
    \includegraphics[width=.95\linewidth]{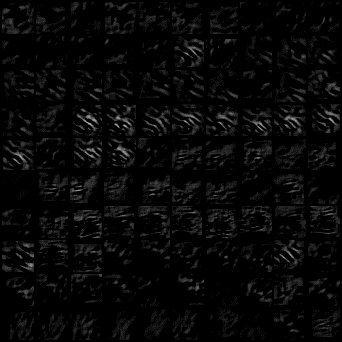}
    Data Impression.
\end{minipage}

\clearpage
\newpage
Joint CIFAR10:
\centering
\begin{minipage}{.5\textwidth}
    \centering
    \includegraphics[width=.95\linewidth]{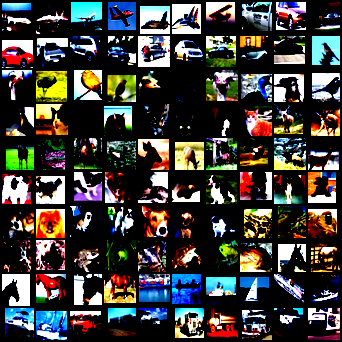}
    Original data.
\end{minipage}%
\begin{minipage}{.5\textwidth}
    \centering
    \includegraphics[width=.95\linewidth]{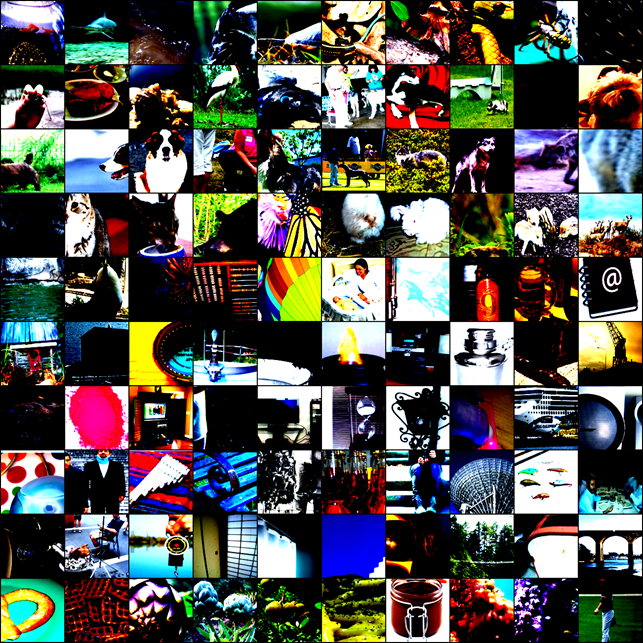}
    Auxiliary data (ImageNet).
\end{minipage}
\begin{minipage}{.5\textwidth}
    \centering
    \includegraphics[width=.95\linewidth]{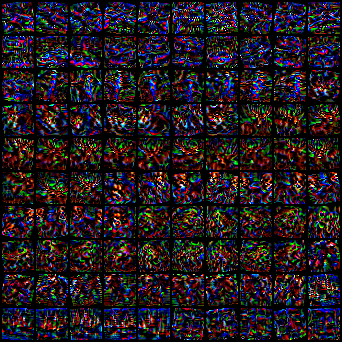}
    Model Inversion.        
\end{minipage}%
\begin{minipage}{.5\textwidth}
    \centering
    \includegraphics[width=.95\linewidth]{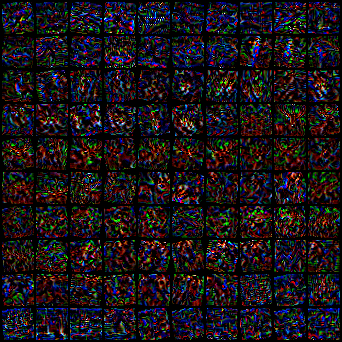}
    Data Impression.
\end{minipage}

\clearpage
\newpage
Split CIFAR10:
\centering
\begin{minipage}{.5\textwidth}
    \centering
    \includegraphics[width=.95\linewidth]{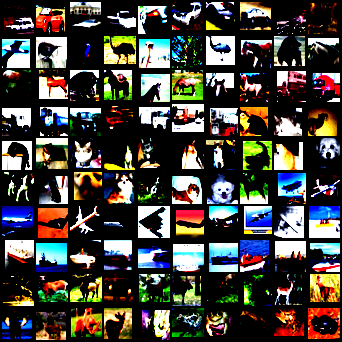}
    Original data.
\end{minipage}%
\begin{minipage}{.5\textwidth}
    \centering
    \includegraphics[width=.95\linewidth]{img/ni_core50/saux.png}
    Auxiliary data (ImageNet).
\end{minipage}
\begin{minipage}{.5\textwidth}
    \centering
    \includegraphics[width=.95\linewidth]{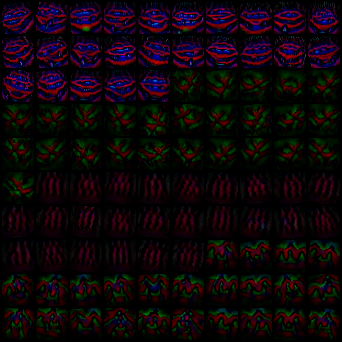}
    Model Inversion.
\end{minipage}%
\begin{minipage}{.5\textwidth}
    \centering
    \includegraphics[width=.95\linewidth]{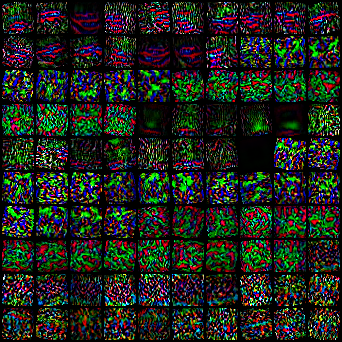}
    Data Impression.
\end{minipage}

\clearpage
\newpage
CORe50-NI:
\centering
\begin{minipage}{.5\textwidth}
    \centering
    \includegraphics[width=.95\linewidth]{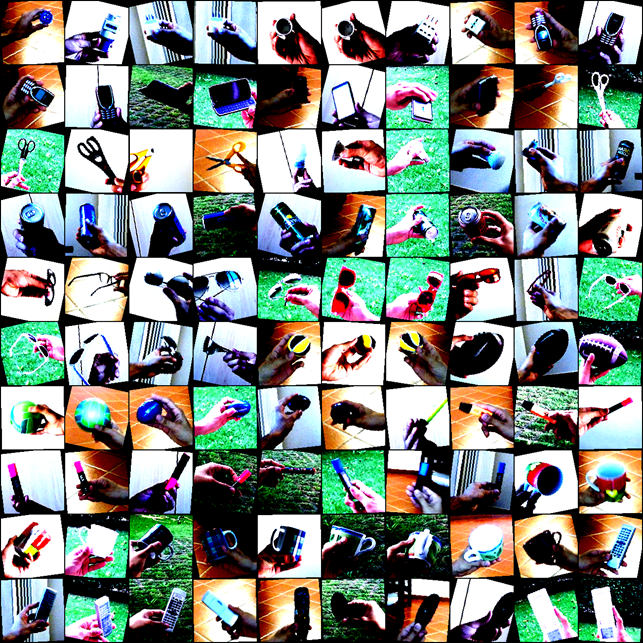}
    Original data.
\end{minipage}%
\begin{minipage}{.5\textwidth}
    \centering
    \includegraphics[width=.95\linewidth]{img/ni_core50/saux.png}
    Auxiliary data (ImageNet).
\end{minipage}
\begin{minipage}{.5\textwidth}
    \centering
    \includegraphics[width=.95\linewidth]{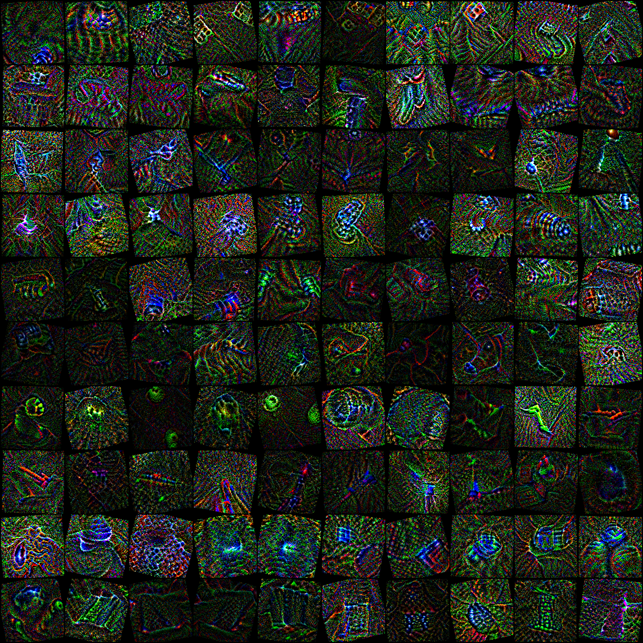}
    Model Inversion.
\end{minipage}%
\begin{minipage}{.5\textwidth}
    \centering
    \includegraphics[width=.95\linewidth]{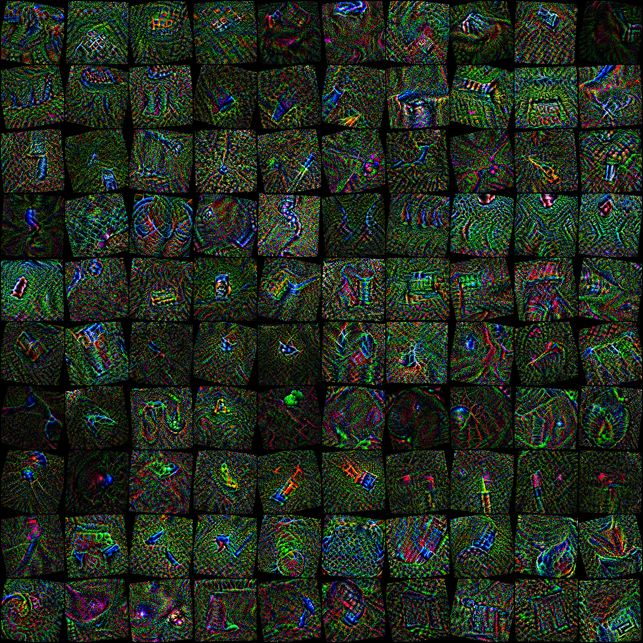}
    Data Impression.
\end{minipage}

\end{document}